\documentclass[twocolumn]{article}
\PassOptionsToPackage{table}{xcolor}
\usepackage[utf8]{inputenc}
\usepackage{times}
\usepackage{newtxmath}
\usepackage{helvet}
\usepackage{courier}
\usepackage{amsmath}
\usepackage{graphicx}
\usepackage{stfloats}
\usepackage{makecell}
\usepackage{adjustbox}
\usepackage[most]{tcolorbox}
\usepackage{listingsutf8}                      %% *** FIX: allow UTF-8 in listings
\tcbset{
  colback=gray!7,
  colframe=black,
  boxrule=0.5pt,
  arc=2pt,
  outer arc=2pt,
  left=6pt,right=6pt,top=6pt,bottom=6pt,
  fonttitle=\bfseries,
  breakable,
  enhanced jigsaw
}

\usepackage[
    letterpaper,
    top=1in,
    bottom=1in,
    left=0.75in,
    right=0.75in,
    headheight=13.6pt
]{geometry}
\usepackage{fancyhdr}

\usepackage{xcolor}
\usepackage{booktabs}
\usepackage{tabularx}
\usepackage{multirow}
\usepackage{soul}

\usepackage{float}
\usepackage[section]{placeins}
\usepackage{caption}
\usepackage{titlesec}
\titleformat{\section}{\normalfont\large\bfseries}{\thesection.}{0.5em}{}
\titleformat{\subsection}{\normalfont\large\bfseries}{\thesubsection.}{0.5em}{}
\titleformat{\subsubsection}{\normalfont\normalsize\bfseries}{\thesubsubsection.}{0.5em}{}
\titlespacing*{\section}{0pt}{3.5ex plus 1ex minus .2ex}{2.3ex plus .2ex}
\titlespacing*{\subsection}{0pt}{3.25ex plus 1ex minus .2ex}{1.5ex plus .2ex}

\usepackage{enumitem}
\setlist[itemize]{itemsep=0pt, topsep=0.pt}

\begin{document}

% --- TITLE AND ABSTRACT (spanning both columns) ---
\twocolumn[
  \begin{@twocolumnfalse} % Temporarily disable two-column mode for the title block
    \title{%
      \vspace{-1em}%
      \parbox{\textwidth}{\centering
        \fontsize{20pt}{24pt}\selectfont\textbf{R-ConstraintBench:}\\[0.25em]
        \fontsize{20pt}{24pt}\selectfont\textbf{Evaluating LLMs on NP-Complete Scheduling}%
      }%
      \vspace{1em}%
    }
   \author{%
      \centering
      \begin{tabular}{@{}c@{\hspace{1.75em}}c@{}}
        % <-- CHANGED: use mini tabulars with [t] so they top-align
        \begin{tabular}[t]{@{}c@{}}
          {\fontsize{12pt}{14pt}\selectfont Raj Jain}\\
          {\fontsize{10pt}{12pt}\selectfont rjain@labelbox.com}
        \end{tabular} &
        \begin{tabular}[t]{@{}c@{}}
          {\fontsize{12pt}{14pt}\selectfont Marc Wetter}\\
          {\fontsize{10pt}{12pt}\selectfont mwetter@labelbox.com}
        \end{tabular} \\
        [2.9em]
        \multicolumn{2}{c}{\fontsize{11pt}{13pt}\selectfont Labelbox}
      \end{tabular}
    }
    \date{} % Clear the date
    \maketitle
    \thispagestyle{empty} % No page number on the first page
    \vspace{-1em}%
    \begin{abstract}
    \noindent\parbox{0.9\textwidth}{%
    Effective scheduling under tight resource, timing, and operational constraints underpins large-scale planning across sectors such as capital projects, manufacturing, logistics, and IT fleet transitions. 
    However, the reliability of large language models (LLMs) when reasoning under high-constraint regimes is insufficiently characterized. 
    To address this gap, we present R-ConstraintBench, a scalable framework that evaluates models on Resource-Constrained Project Scheduling Problems (RCPSP), an NP-Complete feasibility class, while difficulty increases via linear growth in constraints. %% *** EDIT (terminology consistency; concision)
    R-ConstraintBench incrementally increases non-redundant precedence constraints in Directed Acyclic Graphs (DAGs) and then introduces downtime, temporal windows, and disjunctive constraints. %% *** EDIT (parallelism, flow)
    As an illustrative example, we instantiate the benchmark in a data center migration setting and evaluate multiple LLMs using feasibility and error analysis, identifying degradation thresholds and constraint types most associated with failure. %% ***
    Empirically, strong models are near-ceiling on precedence-only DAGs, but feasibility performance collapses when downtime, temporal windows, and disjunctive constraints interact—implicating \emph{constraint interaction}, not graph depth, as the principal bottleneck. Performance on clean synthetic ramps also does not guarantee transfer to domain-grounded scenarios, underscoring limited generalization.

    }
    \end{abstract}
    \vspace{2em}
  \end{@twocolumnfalse}
]

% The rest of your document content follows here.
% The text will now flow in two columns.

\section{INTRODUCTION}

\subsection{Motivation}
Accurate scheduling under resource, timing, and operational constraints is critical across construction, manufacturing turnarounds, airline and crew planning, supply-chain coordination, clinical trial operations, satellite tasking, and large IT transitions. 
As organizations explore LLMs for planning and coordination, small reliability lapses can propagate into cost, safety, and service risks. As a result, ensuring that LLM-generated schedules are feasible is therefore a prerequisite for deployment. Beyond feasibility, systematic evaluation is needed to understand the precise conditions under which LLMs remain reliable, where they break down, and when alternative solution strategies should be employed. To isolate models’ core latent reasoning ability on this task, our evaluation does not permit specialized tool use; tool-augmented pipelines are left to future work.

\vspace{-0.25em}
\subsection{LLM Reliability on Complex RCPSPs}
LLMs show competence on small scheduling puzzles and short dependency chains~\cite{planbench}, but generalization to regimes with interacting precedence, resource, and temporal rules is unclear. We adopt RCPSP feasibility as a focused testbed because it concentrates the reasoning burden on satisfying heterogeneous constraints.
We ask whether models scale their reasoning to deep, structured hierarchies with resource calendars and mutual-exclusion requirements, conditions common in multi-site data center migrations, capital builds, and airline crew pairing.

\vspace{-0.5em}
\subsection{Our Contributions}
In this work, we develop R-ConstraintBench (Resource-Constraint Bench), a controlled benchmarking framework that reveals when and why leading LLMs succeed or fail on RCPSP feasibility tasks. Our key contributions are:
\begin{itemize}
  \item \textbf{Synthetic Constraint Progression}, via a layered DAG generator that incrementally increases non-redundant precedence constraints by one per level, followed by resource downtimes, temporal windows, and disjunctive exclusivity at controlled rates.
  \item \textbf{Domain Instantiation (Illustrative)}, mapping the same RCPSP structure to a data center migration scenario, preserving layered dependencies while placing tasks into representative operational phases and constraints.
  \item \textbf{Comprehensive Evaluation and Infeasibility Analysis}, assessing feasibility and failure modes while identifying breakpoints where reliability degrades.
\end{itemize}
\vspace{-0.5em}

\section{RELATED WORK}
Recent efforts connect LLMs with operations research (OR) by training models to translate natural language into optimization models. OR-Instruct develops a semi-automated data synthesis pipeline and an IndustryOR benchmark to train ORLMs for optimization modeling, with competitive results on NL4OPT~\cite{nl4opt} and MAMO when models are finetuned for OR tasks~\cite{or-instruct}. OptiMUS similarly targets scalable modeling via LLMs, emphasizing end-to-end pipelines that produce executable code for MIP/CP solvers~\cite{optimus, lm4opt}. MAMO frames mathematical modeling as a benchmark tied to solver correctness, moving evaluation beyond free-form text to executable outputs~\cite{mamo}. Our scope differs: rather than broad modeling translation, R-ConstraintBench isolates feasibility on NP-Complete RCPSP instances, varies the number of active constraints systematically, and provides an infeasibility analysis exposing failure modes. %% *** EDIT (tighter phrasing)

Work on \emph{LLMs for Scheduling} shows that models can be prompted to generate valid CP models for Job-Shop Scheduling and solve them with CP solvers~\cite{llms-can-schedule}. Starjob contributes a modern dataset tailored to LLM-driven JSSP, enabling finetuning and prompting comparisons~\cite{starjob}. By contrast, our benchmark does not require models to construct CP-SAT solvers; the focus of this study is to understand inherent reasoning ability rather than optimizing for a particular task where tool use would be employed.

A complementary thread studies \emph{text-to-formalism and verification-in-the-loop for planning and constraint reasoning}~\cite{cfbench}. Planetarium evaluates translation from natural language into planning languages~\cite{planetarium}. Other work augments LLM outputs with satisfiability/SMT solvers to catch inconsistencies in decision pipelines~\cite{verification_smt}. Relatedly, \emph{streamlining constraints} use auxiliary constraints to ease solving without sacrificing correctness~\cite{streamlining_constraints}. R-ConstraintBench keeps a formal verifier in the loop but focuses on RCPSP feasibility and structured failure attribution.

\section{PROBLEM FORMULATION}
We formalize RCPSP feasibility and the axes of complexity we vary.

\vspace{-0.5em}
\subsection{RCPSP as a Feasibility Problem}

A Resource-Constrained Project Scheduling Problem (RCPSP) asks: ``Given a set of tasks, resources, and constraints, can we assign start and end times such that the entire schedule is valid?'' In our formulation, we do not minimize makespan; instead, we require \emph{complete} constraint satisfaction. This choice isolates feasibility reasoning from optimization objectives, enabling binary verification and attributing errors to constraint handling rather than objective trade-offs, consistent with our focus on latent reasoning ability and comparable difficulty across instances as constraint density increases.
Formally, an RCPSP instance comprises a collection of tasks \(T=\{1,\dots,n\}\), each with fixed duration \(p_i\); a set of resources \(R=\{1,\dots,m\}\), each with time-varying capacity \(c_r(t)\); resource demands \(q_{i,r}\) denoting how much of resource \(r\) task \(i\) consumes; and precedence relations \(i\to j\) that forbid task \(j\) from starting before task \(i\) completes.

\noindent A schedule assigns start times \(s_i \ge 0\) so that:
\vspace{0.1em}

\noindent\textbf{Precedence:}
\begin{equation}
  s_j \ge s_i + p_i,\quad \forall\,(i\to j).
\end{equation}

\noindent\textbf{Resource feasibility:}
\begin{equation}
\sum_{i:\,s_i \le t < s_i + p_i} q_{i,r}
  \le c_r(t),\quad \forall\, r\in R,\ \forall\, t\ge 0.
\end{equation}

\FloatBarrier
\begin{figure}[!t]
  \centering
  \includegraphics[width=\columnwidth]{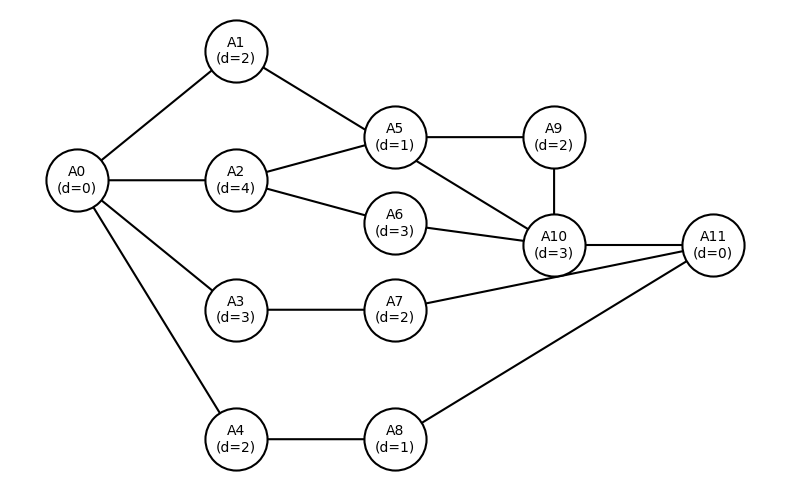}
  \caption{Precedence activity with duration on a node graph}
  \label{fig:rcpsp}
\end{figure}
\FloatBarrier

\vspace{-0.5em}
\subsection{Representing RCPSP as DAGs}

Any set of precedence constraints prohibiting cycles forms a Directed Acyclic Graph (DAG). Beyond tautology, using layered DAGs to represent RCPSPs gives three key benefits:
\begin{enumerate}
  \item \textbf{Natural modeling of phases.} Large programs (e.g., construction sequences, release trains, data center relocations) decompose into stages; layers mirror these stages.
  \item \textbf{Guaranteed acyclicity and clarity.} Tasks with no circular dependencies are fundamental to feasibility; DAG structure enforces this at origin. Layers prevent inadvertent back-edges and clearly separate “earlier” from “later” tasks.
  \item \textbf{Controlled complexity progression.} We add one new cross-layer precedence edge at a time; at level \(k\), the DAG has exactly \(k\) non-redundant constraints, isolating the effect of increasing dependency depth.
\end{enumerate}
\vspace{-0.5em}

\subsection{DAG Generation}

To operationalize layered generation over \(m\) layers, we partition tasks into layers \(L_1, L_2, \dots, L_m\) and initialize an empty reachability matrix. Then, for each of \(k\) desired precedence-constraint levels, we sample a candidate edge \((i, j)\) where \(i \in L_a\) and \(j \in L_b\) with \(b > a\). We accept this edge only if no path \(i \to j\) already exists in the transitive closure, and subsequently update the reachability matrix so that later levels cannot re-introduce any implied edges. This guarantees acyclic, non-redundant structures. Parameters \(m\) and \(k\) allow fine-grained control over the DAG’s hierarchy and density.

\vspace{-0.5em}
\subsection{Realism via Additional Constraint Axes}

A pure-precedence DAG is the skeleton. We layer on three more constraint types:
\begin{itemize}
  \item \textbf{Resource Downtime:} Certain resources become unavailable (\(c_r(t)=0\)).
  \item \textbf{Temporal Windows:} Release times \(r_i\) (\(s_i\ge r_i\)); deadlines \(d_i\) (\(s_i+p_i\le d_i\)).
  \item \textbf{Disjunctive (No-Overlap):} For some \((i,j)\), enforce \(s_i+p_i\le s_j\) or \(s_j+p_j\le s_i\).
\end{itemize}
\vspace{0.25em}

Each axis is sampled at controlled probabilities (e.g., 75\%). The result is a parameterized, multi-axis challenge suitable for systematic evaluation, forming the core of our R-ConstraintBench evaluation.
\section{SYSTEM DESIGN AND EXPERIMENTAL METHODOLOGY}

R-ConstraintBench’s evaluation pipeline consists of three integrated components:
(1) structured instance generation with guaranteed solvability,
(2) controlled LLM scheduling, and
(3) exact automated verification of each candidate schedule.
Separating generation, model interaction, and validation helps ensure measured performance reflects model reasoning rather than data noise or human intervention. 

\subsection{Structured Instance Generation}

We synthesize RCPSP problem instances that are realistic and tunable in difficulty. Leveraging an RCPSP solver program (built on OR-Tools’ CP-SAT), we enforce a two-step process:

\begin{enumerate}
  \item \textbf{Layered-DAG Construction.}

  Partition tasks into \(m\) layers (e.g., 2--5); initialize reachability. For each of \(k\) levels, add exactly one cross-layer edge if it is non-redundant, updating reachability immediately. The result is a DAG with precisely \(k\) non-redundant dependencies.
  
\vspace{0.5em}
  \item \textbf{Constraint Injection and Feasibility Check.}

  Randomly assign each task a duration drawn from \([1, D_{\max}]\), ensure every task demands at least one of three resource types, impose temporal windows on a subset of tasks, introduce downtime intervals during which certain resources are fully unavailable, and add disjunctive constraints forbidding simultaneous execution of specific task pairs.

  Immediately after generation, we invoke the solver to check feasibility. Any instance that fails to admit a valid schedule is discarded and regenerated, guaranteeing that every instance entering evaluation admits a feasible schedule.
\end{enumerate}

\subsection{Controlled LLM Scheduling}

Once instances are generated, we evaluate each LLM under identical conditions:
\begin{itemize}
  \item \textbf{Prompt Design.} A concise natural-language description lists every task’s duration, resource needs, and all constraints, followed by a JSON-schema template defining the expected ‘schedule’ output format (Appendix). To keep evaluation focused on inherent model capability rather than prompt-engineering technique, we adopt a single-shot format that requires the model to integrate all constraints and produce a complete schedule in one pass~\cite{cot,tot,react,pal}.
  \item \textbf{Response Capture.} We test a representative suite of models. Each instance is presented in a single-shot prompt, and model outputs are parsed into structured candidate schedules. This setup requires the model to interpret the entire scheduling problem holistically, integrating all constraints into a coherent plan without intermediate guidance or iterative feedback loops.
\end{itemize}

\FloatBarrier
\begin{figure}[!t]
  \centering
  \includegraphics[width=\columnwidth]{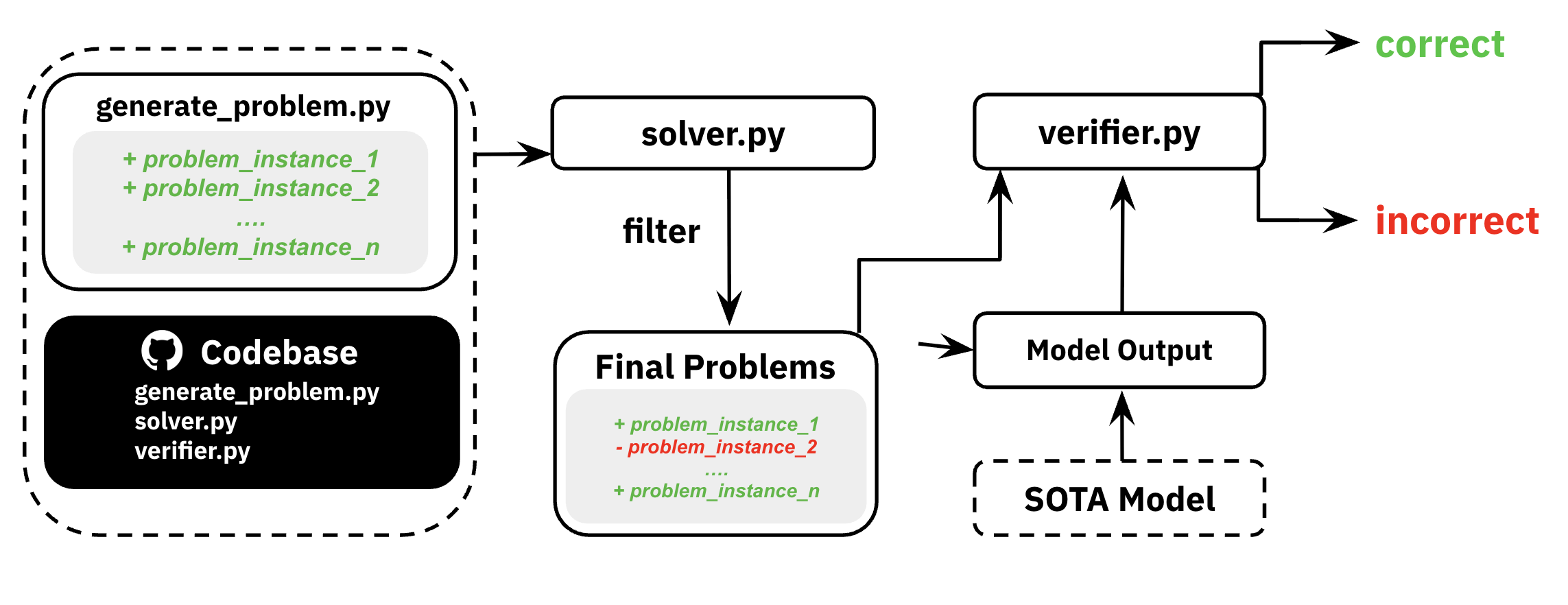}
  \caption{An overview of R-ConstraintBench system design}
  \label{fig:system-design}
\end{figure}
\FloatBarrier

\subsection{Exact Automated Verification}

To measure correctness, every candidate schedule is fed into a verifier program, which checks:
\begin{itemize}
  \item \textbf{Precedence Compliance.} No task starts before all its predecessors finish.
  \item \textbf{Resource and Downtime Adherence.} At each time step, resource usage never exceeds capacity, and resources remain idle during downtime windows.
  \item \textbf{Temporal Window Satisfaction.} All start times respect release dates and deadlines.
  \item \textbf{Disjunctive Enforcement.} For each forbidden pair \((i,j)\), tasks do not overlap in time.
\end{itemize}

Violations are automatically classified by type (precedence, resource/downtime, temporal, disjunctive), enabling a fine-grained infeasibility analysis. Together, this three-stage design—generation, scheduling, verification—forms the backbone of R-ConstraintBench, providing a controlled evaluation of LLM reliability on increasingly complex RCPSP feasibility tasks.

\section{BENCHMARK CONSTRUCTION}

To rigorously evaluate LLM scheduling reliability, we assembled a suite of synthetic RCPSP benchmarks whose structure and operational complexity can be dialed. Every instance is generated by our solver-driven pipeline (Section 5) and passed through a CP-SAT solver to guarantee feasibility before model evaluation. We organize our evaluation into two phases -- first isolating pure precedence reasoning, then layering on full operational constraints -- culminating in an illustrative data center migration scenario.

\subsection{Phase I: Pure-Precedence DAGs}

Phase I strips the problem to its dependency graph, so any model failure stems from mis-reasoning about precedence. Because Phase I is pure-precedence DAGs, the identification of any valid topological sort is admitted as a solution. Finding a topological sort of a precedence DAG can be done in linear time \(O(V+E)\). Therefore, these are not NP-Complete problems.

\begin{enumerate}
  \item \textbf{Two-Layer Bipartite DAGs:}
    \begin{enumerate}
      \item Levels 1--200: level \(k\) has exactly \(k\) edges.
      \item Zero transitive/redundant edges.
      \item 10 instances per level \(\to\) 2,000 total.
    \end{enumerate}
    \vspace{0.1em}

  \item \textbf{Multi-Layer DAGs:}
    \begin{enumerate}
      \item Layer counts: 3-, 4-, and 5-layer configurations.
      \item Levels 1--200: exactly \(k\) non-redundant cross-layer edges at level \(k\).
      \item 10 instances per level, per layer count \(\to\) \(2{,}000\times3=6{,}000\) total.
    \end{enumerate}
\end{enumerate}

\subsection{Phase IIa: Multi-Constraint Interaction}

Phase IIa fixes a 5-layer backbone and injects three additional constraint types at 75\% probability to emulate the structure of complex, NP-Complete RCPSP problems. The 75\% probability was selected to balance realism and difficulty — high enough to meaningfully stress models with frequent constraint interactions, yet low enough to avoid near-unsolvable instances at modest precedence levels. These cannot be solved in linear time because checking feasibility is no longer a pure ordering problem.

\begin{enumerate}
  \item \textbf{5-Layer DAGs:} Levels 1--200 each feature exactly \(k\) non-redundant left \(\to\) right edges (level \(k\) has \(k\) edges) with zero transitive dependencies. Temporal, downtime, and disjunctive constraints are each applied independently at 75\% probability, and 10 instances are generated per level—2,000 total.
\end{enumerate}

\subsection{Phase IIb: Data Center Migration}
\vspace{0.5em}
We map Phase IIa (75/75/75) onto Phase IIb, an illustrative data center migration narrative:

\begin{itemize}
  \item \textbf{Five-stage rack workflow:} Each server rack follows Shutdown \(\to\) Unrack \(\to\) Transport \(\to\) Install \(\to\) Test, mirroring the 5-layer DAG backbone.
  \item \textbf{Specialized Resources:} \texttt{IT\_Team}, \texttt{DC\_Crew}, \texttt{Network\_Engineers}, \texttt{Forklift} and \texttt{Convoy} appear with real capacities (e.g.\ single-Forklift, 3-person crew).
  \item \textbf{Downtime windows (75\%):} Periods during which specific resources are entirely unavailable (e.g., network maintenance, forklift inspection) are introduced.
  \item \textbf{Temporal windows (75\%):} Key steps carry earliest-start or deadline constraints to model regulatory cut-over windows.
  \item \textbf{Disjunctive conflicts (75\%):} Tasks sharing unique assets (e.g.\ Forklift) cannot overlap, enforcing mutual exclusion.
\end{itemize}

10 instances are generated per level—2,000 total. By mapping exactly \(k\) non-redundant cross-rack precedence plus these three constraint axes onto a rack migration, we obtain RCPSP instances that both reflect operational complexity and retain the NP-Complete hardness of Phase II. Phase II datasets are used to calculate model performance on R-ConstraintBench.

\section{EVALUATION SETUP}
\subsection{Research Questions}

We evaluate the performance of state-of-the-art (SOTA) LLMs on RCPSP problem solving with R-ConstraintBench. Specifically, we aim to address the following research questions (RQs):

\begin{enumerate}
  \item \textbf{RQ1: Constraint Scalability.} At what level of added precedence constraints does each model’s feasibility rate drop below a usable threshold?
  \item \textbf{RQ2: Constraint-Type Sensitivity.} How do resource downtime, temporal windows, and disjunctive constraints collectively impact model reliability when layered onto a fixed DAG?
  \item \textbf{RQ3: Domain Transferability.} Can performance patterns observed on synthetic benchmarks predict reliability in an illustrative, business scenario?
\end{enumerate}

\FloatBarrier
\begin{figure*}[!t]
  \centering
  \includegraphics[height=0.5\textheight,keepaspectratio]{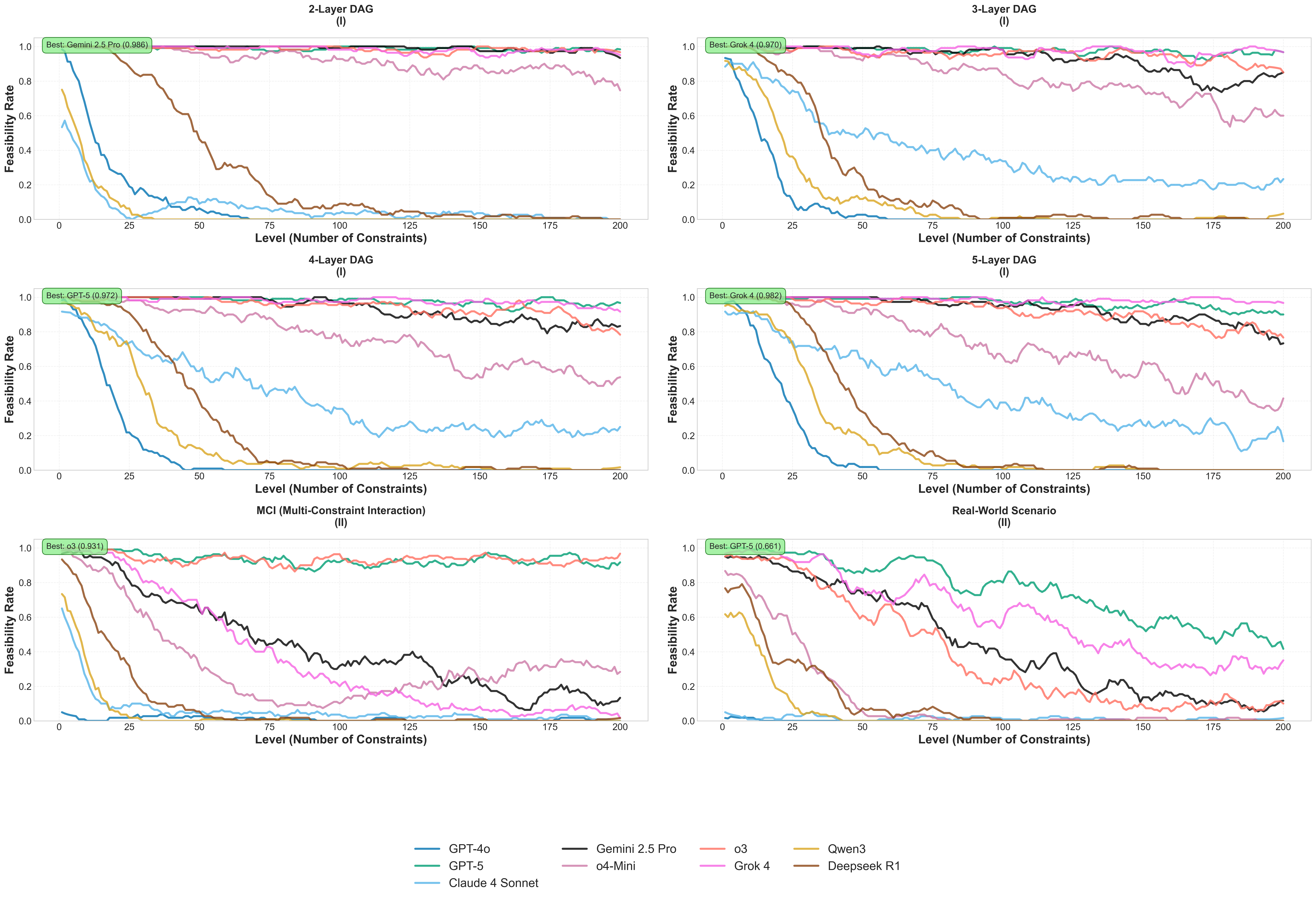}
  \caption{Smoothed feasibility rates vs. constraint levels using 10-point moving average windows}
  \label{fig:feasibility-curves}
\end{figure*}

\subsection{Model Selection}

Solving an RCPSP instance in R-ConstraintBench demands deep combinatorial reasoning, the ability to juggle long contextual descriptions, and careful handling of heterogeneous constraint types. To probe these capabilities, we select nine SOTA LLMs for evaluation. Two of them are open-source models (DeepSeek-R1-0528 and Qwen3-235B-A22B, where DeepSeek-R1 is a reasoning model and Qwen3-235B-A22B is used in “thinking” mode). The other seven are closed-source models (Claude-4-Sonnet, GPT-4o-2024-11-20, Gemini-2.5-Pro, o4-mini-2025-04-16, o3-pro-2025-06-10, GPT-5-2025-08-07, and Grok-4-0709). For brevity, version numbers will be omitted when referencing model names throughout the paper.

\subsection{Evaluation Metrics}

We measure model performance along three complementary dimensions—accuracy, robustness, and failure‐mode breakdown. All metrics are computed per model \(M\) and per dataset \(D\), where each dataset spans \(K\) levels (e.g., \(K{=}200\)) with 10 instances per level (\(k=1,\dots,K\)).

\begin{enumerate}
  \item \textbf{Feasibility Rate (Feasibility \%)}:
  For each level \(k\),
  \begin{equation}
    F^M_D(k) = \frac{\#\{\text{instances at level }k\text{ solved correctly}\}}{10}\,, \tag{3}
  \end{equation}
  i.e., the fraction of the 10 instances at level \(k\) that the model schedules feasibly.
  \vspace{0.5em}

  \item \textbf{Weighted Area Under the Curve (WAUC)}:
  Because difficulty increases monotonically with \(k\) (one additional constraint per level), later levels are weighted more heavily via linear weights
  \begin{equation}
    w_k \;=\; \frac{2k}{K(K{+}1)} \quad \text{with} \quad \sum_{k=1}^{K} w_k \;=\; 1\,. \tag{4}
  \end{equation}
  The dataset‐level robustness is then
  \begin{equation}
    \mathrm{WAUC}^M_D \;=\; \sum_{k=1}^{K} w_k\, F^M_D(k)\,, \tag{5}
  \end{equation}
  which rewards models that remain feasible at higher constraint levels.
  \vspace{0.5em}

  \item \textbf{Breakpoint Level (BP‐Level)}:
  The smallest level \(k\) at which
  \begin{equation}
    F^M_D(k) \;<\; \tau \,, \tag{6}
  \end{equation}
  for a high‐accuracy threshold \(\tau\) (e.g., \(70\%\)).
  \vspace{0.5em}

  \item \textbf{Infeasibility Analysis}:
  Every failed schedule is automatically classified by the verifier into one or more of four violation categories: \emph{Precedence}, \emph{Resource/Downtime}, \emph{Temporal}, and/or \emph{Disjunctive}.
\end{enumerate}

\noindent\textbf{Official R-ConstraintBench Score.}
Benchmark’s headline score for model \(M\) averages WAUC over the Phase-II datasets:
\begin{equation}
  \mathrm{Score}^M \;=\; \frac{1}{\lvert \mathcal{D}_{\mathrm{phase2}} \rvert} \sum_{D \in \mathcal{D}_{\mathrm{phase2}}} \mathrm{WAUC}^M_D \,.
  \tag{7}
\end{equation}
This definition aligns the score with practical difficulty, emphasizing high-constraint regimes while remaining comparable across datasets.

\section{RESULTS}
We report results across the full 200-level range using per-level feasibility (Fig.~3), weighted area under the curve (WAUC; Fig.~4), breakpoints at \(\tau{=}70\%\) (Table~1), and per-dataset comparisons including the real-world domain (Fig.~5, Table~2).
We set \(\tau{=}70\%\) to reflect an operational-majority criterion—requiring at least 7 of 10 instances per level to be feasible—which is stringent enough to flag fragile behavior while avoiding over-penalization for occasional parsing or formatting errors.

\subsection{Phase I: Pure-Precedence Baselines}
Across two- to five-layer DAGs, top models sustain high feasibility across nearly all levels (Fig.~3), reflected in near-ceiling WAUCs (Fig.~4). \textbf{Grok~4} is strongest overall on pure precedence (Phase~I WAUC \(=0.975\)), followed closely by \textbf{GPT-5} \((0.971)\), \textbf{o3} \((0.933)\), and \textbf{Gemini~2.5~Pro} \((0.919)\). Mid-tier models trail (\textbf{o4-Mini}: \(0.732\)), and smaller/general models are far below (Table~2). Breakpoints mirror this pattern (Table~1): for example, Grok~4, GPT-5, and o3 exhibit \(>\!200\) on 2-, 3-, 4-, and 5-layer datasets. These baselines confirm that certain SOTA LLMs can reliably produce feasible schedules when constraints are purely precedence-based.

\subsection{Phase IIa: Multi-Constraint Interaction}
Introducing downtime, temporal windows, and disjunctive no-overlap constraints (\textbf{Multi-Constraint Interaction; MCI}) sharply separates models (Fig.~3, Fig.~4). \textbf{o3} dominates the synthetic MCI constraints set with \(\textbf{WAUC}=0.931\), with \textbf{GPT-5} competitive just behind the leader \((0.924)\) and above both \textbf{Gemini~2.5~Pro} \((0.289)\) and \textbf{Grok~4} \((0.183)\) (Table~2). Breakpoints align with this picture (Table~1): \textbf{o3} and \textbf{GPT-5} never drop below \(\tau\) across the full 200-level ramp (\(>\!200\)), while \textbf{Grok~4} and \textbf{Gemini} cross at 40 and 50 respectively; \textbf{o4-Mini} crosses at 25, and the remaining models much earlier. For weaker systems, feasibility collapses soon after constraint interactions become frequent (levels \(\approx 10\text{--}30\)), whereas \textbf{o3} and \textbf{GPT-5} maintain high feasibility deep into the ramp (Fig.~3).

\begin{table}[!t]
  \centering
  \label{tab:breakpoints}
  \resizebox{\columnwidth}{!}{
  \begin{tabular}{lrrrrrr}
    \toprule
    \textbf{Model} & \textbf{2-Layer} & \textbf{3-Layer} & \textbf{4-Layer} & \textbf{5-Layer} & \textbf{MCI} & \textbf{Real-World} \\
    \midrule
    % UPDATED VALUES
    Claude 4 Sonnet & 4 & 28 & 32 & 39 & 6 & 1 \\   % <-- updated
    DeepSeek R1     & 40 & 34 & 37 & 41 & 12 & 13 \\ % <-- updated
    GPT-4o          & 10 & 9  & 18 & 18 & 1  & 1 \\  % <-- updated
    GPT-5           & $>200$ & $>200$ & $>200$ & $>200$ & $>200$ & 95 \\ % <-- new row
    Gemini 2.5 Pro  & $>200$ & $>200$ & $>200$ & 199 & 50 & 50 \\ % <-- updated
    Grok 4          & $>200$ & $>200$ & $>200$ & $>200$ & 40 & 61 \\ % <-- updated
    Qwen3           & 6  & 18 & 17 & 22 & 6  & 4 \\   % <-- updated
    o3              & $>200$ & $>200$ & $>200$ & $>200$ & $>200$ & 49 \\ % <-- updated
    o4-Mini         & 195 & 120 & 129 & 100 & 25 & 16 \\ % <-- updated
    \bottomrule
  \end{tabular}}
  \caption{Breakpoint level (BP; lowest \(k\) where Feasibility\% \(<\tau\)) by model and dataset.}
\end{table}

\subsection{Phase IIb: Data Center Migration}
On the data center migration domain, the top of the ranking changes (Fig.~4). \textbf{GPT-5} achieves the best operational performance with \(\textbf{WAUC}=0.661\) and the latest real-world breakpoint (BP \(=\) 95), ahead of \textbf{Grok~4} (WAUC \(=\) 0.483, BP \(=\) 61) and \textbf{Gemini~2.5~Pro} (WAUC \(=\) 0.282, BP \(=\) 50); \textbf{o3} drops to \(\textbf{0.214}\) with BP \(=\) 49 (Table~2). These results indicate that domain-specific coupling (e.g., single-capacity equipment overlapping with maintenance windows and deadlines) alters the failure surface in ways not fully captured by synthetic MCI instances.

\subsection{R-ConstraintBench Score}
The benchmark’s headline score averages Phase-II WAUC across the synthetic MCI constraints set and the data center migration set (Sec.~4.3). The top tier consists of \textbf{GPT-5} and \textbf{o3}, followed by \textbf{Grok~4} and \textbf{Gemini~2.5~Pro} (Table~2). \textbf{GPT-5} attains the highest overall score \textbf{(0.792)} due to its leading real-world WAUC (0.661) and competitive synthetic standing. Together with breakpoints (Table~1) and feasibility curves (Fig.~3), these results delineate the current reliability frontier under increasing constraint density and domain realism.

\section{ANALYSIS}

\subsection{Structure vs.\ Constraint Interaction}
Phase~I shows that the strongest models (GPT-5, Grok~4, o3, Gemini~2.5~Pro) reliably handle precedence-only scheduling. Phase~II reveals, however, that \emph{constraint interaction}---overlapping downtime windows, deadlines, and disjunctive pairs---rapidly erodes feasibility for most models (Fig.~3, Fig.~4). The dispersion in breakpoints (Table~1) suggests that “graph reasoning” competence is necessary but insufficient for \emph{resource--temporal coupling}, which demands globally consistent reasoning across heterogeneous rules.

\begin{figure*}[t]
  \centering
  \includegraphics[width=\textwidth,height=0.3\textheight,keepaspectratio]{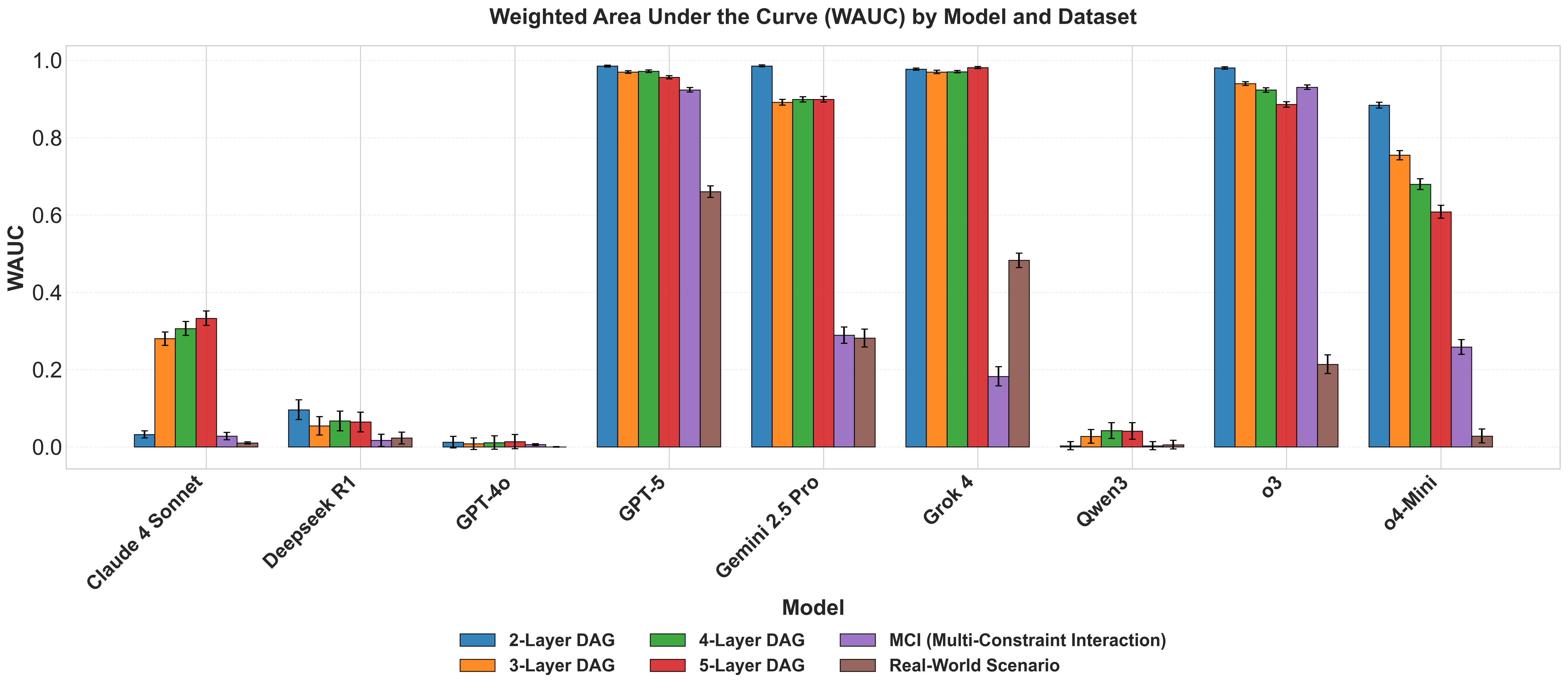}
  \captionsetup{justification=centering}
  \caption{WAUC calculations vs. model per dataset.}
  \label{fig:wauc-comparison}
\end{figure*}

\begin{table*}[!t]
    \centering
    \label{tab:final_data_swapped}
    
    \footnotesize 
    \setlength{\tabcolsep}{4pt} 

    \begin{tabularx}{\textwidth}{l *{6}{>{\centering\arraybackslash}X} >{\centering\bfseries\arraybackslash}X *{3}{>{\centering\arraybackslash}X}}
        \toprule
        \textbf{Model} & 
        \textbf{PI WAUC} & 
        \textbf{PI AUC} & 
        \textbf{PI Succ.\%} & 
        \textbf{PII WAUC} & 
        \textbf{PII AUC} & 
        \textbf{PII Succ.\%} & 
        \textbf{Bench. Score} & 
        \textbf{5-L WAUC} & % <-- HEADER CHANGED
        \textbf{MCI \newline WAUC} & 
        \textbf{Real-World\newline WAUC} \\
        \midrule
         % NEW ROW ADDED
        GPT-5           & 0.971 & 0.979 & 97.9\% & 0.792 & 0.847 & 84.7\% & 0.792 & 0.956 & 0.924 & 0.661 \\ % <-- inserted
        o3              & 0.933 & 0.952 & 95.2\% & 0.572 & 0.661 & 66.0\% & 0.572 & 0.886 & 0.931 & 0.214 \\
        Grok 4          & 0.975 & 0.981 & 98.1\% & 0.333 & 0.487 & 48.7\% & 0.333 & 0.982 & 0.183 & 0.483 \\
        Gemini 2.5 Pro  & 0.919 & 0.948 & 94.8\% & 0.286 & 0.443 & 44.3\% & 0.286 & 0.899 & 0.289 & 0.282 \\
        o4-Mini         & 0.732 & 0.807 & 80.8\% & 0.144 & 0.231 & 23.1\% & 0.144 & 0.608 & 0.259 & 0.028 \\
        Deepseek R1     & 0.071 & 0.234 & 23.6\% & 0.020 & 0.100 & 10.0\% & 0.020 & 0.064 & 0.017 & 0.023 \\
        Claude 4 Sonnet & 0.238 & 0.334 & 33.4\% & 0.019 & 0.037 & 3.7\%  & 0.019 & 0.333 & 0.028 & 0.010 \\
        Qwen3           & 0.028 & 0.123 & 12.3\% & 0.004 & 0.046 & 4.6\%  & 0.004 & 0.041 & 0.003 & 0.006 \\
        GPT-4o          & 0.011 & 0.088 & 8.8\%  & 0.003 & 0.005 & 0.5\%  & 0.003 & 0.014 & 0.006 & 0.000 \\
        \bottomrule
    \end{tabularx}
    \caption{Complete performance metrics for all models across all benchmark phases and datasets.}
\end{table*}

\subsection{Domain Transfer and the Real-World Gap}

Across models, Phase~II shows that complex, operational RCPSPs remain beyond current LLM reasoning. Several models score highly on the MCI constraints track, yet performance shifts on the data center migration domain in ways dependent on the model (Fig.~5, Table~2). \textbf{GPT-5} attains the best real-world score (\(\mathrm{WAUC}=0.661\)), ahead of \textbf{Grok~4} (\(0.483\)), \textbf{Gemini~2.5~Pro} (\(0.282\)), and \textbf{o3} (\(0.214\)). The gap view in Fig.~5 shows contrasting transfer patterns: \textbf{Grok~4} improves when constraints are embedded in realistic couplings (positive real-world gap), while \textbf{o3} exhibits the largest negative gap; \textbf{GPT-5} also shows a negative gap relative to its synthetic standing but still leads operationally. These shifts indicate that domain-specific context changes the failure surface in ways not captured by synthetic stress alone. Moreover, the real-world feasibility curves (Fig.~3) reveal a pronounced late-level decline for all models—including GPT-5—which remain strong through the mid band and then erode as constraint density approaches the tail. Even when a model leads on average, the upper end of the difficulty ramp is a bottleneck for sustained feasibility.

\subsection{Reliability Thresholds and Practicality}

Taken together—feasibility curves (Fig.~3), breakpoints at \(\tau{=}70\%\) (Table~1), and WAUCs (Table~2)—the evidence is consistent: \emph{even the strongest LLMs struggle to sustain feasibility at the highest difficulty bands}. \textbf{GPT-5} lifts the real-world ceiling (best WAUC \(0.661\), latest real-world BP \(=\) 95) yet still declines before upper levels; \textbf{Grok~4} transfers well but likewise degrades (BP \(=\) 61); \textbf{Gemini~2.5~Pro} sits between extremes (BP \(=\) 50); and \textbf{o3} excels synthetically but suffers the largest operational drop (BP \(=\) 49). The downward trajectories across all curves, notably in the 150–200 band, make the unreliability explicit: every model ultimately bends toward low feasibility as constraints accumulate.

\section{INFEASIBILITY ANALYSIS}
\label{sec:error-taxonomy}

To better understand the nature of model failures, we classify every invalid schedule into violation categories.

\begin{itemize}
    \item \textbf{Precedence}: At least one task begins before all of its prerequisite tasks have completed.
    \item \textbf{Resource/Downtime}: Resource capacities are exceeded at one or more time steps, or tasks are scheduled during periods when the required resource is unavailable.
    \item \textbf{Temporal}: Violations of earliest-start or latest-finish requirements, where a task begins too early or finishes late.
    \item \textbf{Disjunctive}: Two tasks that must not overlap in time (due to exclusivity constraints) are scheduled concurrently.
    \item \textbf{Other}: Failures that do not fall neatly into the above categories, such as malformed outputs that cannot be parsed, conflicting minor violations, or combinations of multiple small errors.
\end{itemize}

\FloatBarrier
\begin{figure*}[!t]
  \centering
    \includegraphics[width=\textwidth,height=0.27\textheight,keepaspectratio]{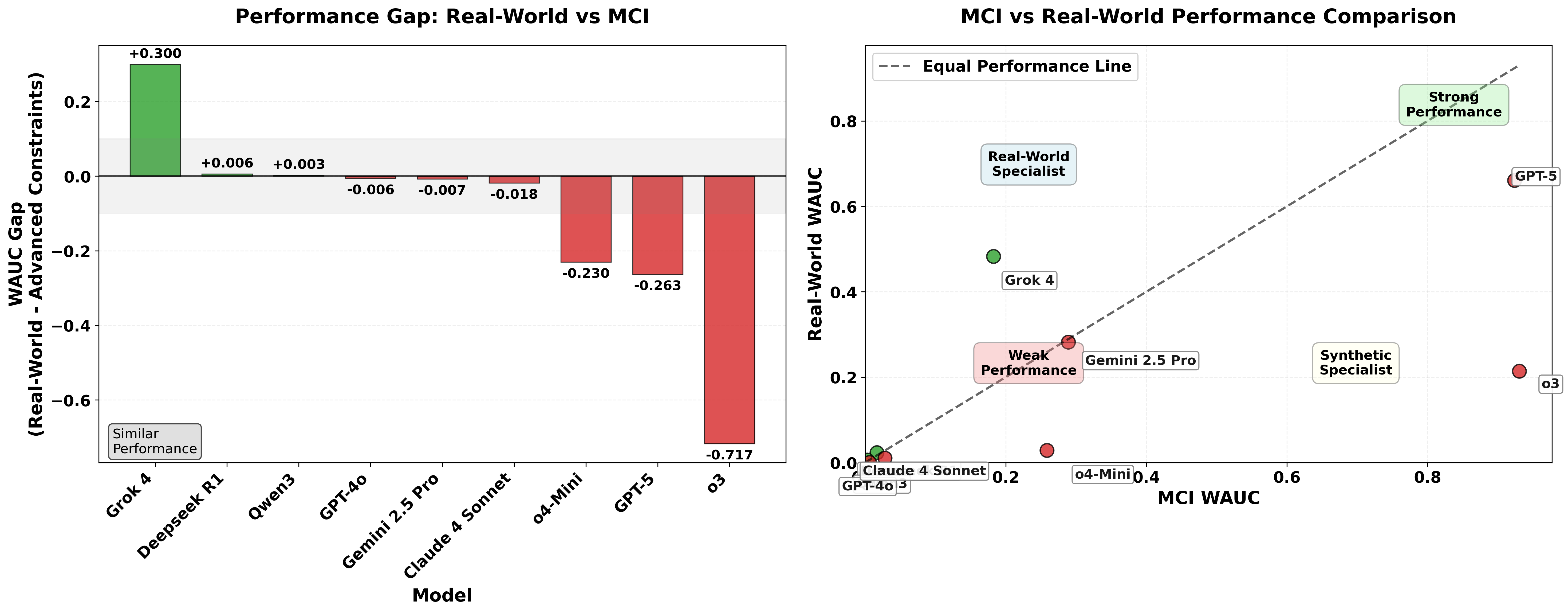}
  % Add this line to center the caption
  \captionsetup{justification=centering} 
  \caption{Performance gap between real world and MCI constraints} 
  \label{fig:gap}
\end{figure*}
\FloatBarrier

\begin{table}[!t]
    \centering
    \label{tab:error_taxonomy}
    \begin{adjustbox}{max width=\columnwidth}
    \begin{tabular}{l r r r r r}
        \toprule
        \textbf{Model} & \textbf{Precedence} & \textbf{\makecell{Resource/\\Downtime}} & \textbf{Temporal} & \textbf{Disjunctive} & \textbf{Other} \\
        \midrule
        Claude 4 Sonnet & 41.6\% & 11.0\% & 12.1\% & 23.2\% & 12.1\% \\
        Deepseek R1     & 41.8\% & 7.6\%  & 7.5\%  & 17.5\% & 25.6\% \\
        GPT-4o          & 64.8\% & 6.5\%  & 11.2\% & 10.2\% & 7.3\%  \\
        GPT-5          & 68.4\% & 9.1\%  & 8.9\% & 7.0\% & 6.6\%  \\
        Gemini 2.5 Pro  & 42.9\% & 6.1\%  & 3.9\%  & 45.4\% & 1.7\%  \\
        Grok 4          & 38.5\% & 8.7\%  & 3.1\%  & 48.4\% & 1.2\%  \\
        Qwen3           & 56.8\% & 11.2\% & 6.8\%  & 19.7\% & 5.4\%  \\
        o3              & 93.7\% & 0.6\%  & 3.7\%  & 1.0\%  & 1.0\%  \\
        o4-Mini         & 70.7\% & 5.2\%  & 4.9\%  & 18.8\% & 0.5\%  \\
        \bottomrule
    \end{tabular}
    \end{adjustbox}
    \caption{Infeasibility percentage breakdown for invalid schedules.}
\end{table}

\subsection{Observed Patterns Across Models}
Table~3 summarizes the percentage breakdown of these violation types for each evaluated model. Several notable patterns emerge:

\begin{itemize}
    \item \textbf{Precedence violations dominate for certain models.} \textbf{o3} exhibits an exceptionally high rate of precedence errors (93.7\%), indicating that once it fails, it almost always does so by misordering tasks. Other models such as \textbf{o4-Mini} (70.7\%), \textbf{GPT-5} (68.4\%) and \textbf{GPT-4o} (64.8\%) also show a strong bias toward precedence failures.
    \item \textbf{Disjunctive constraint violations are a major weakness for Grok 4 and Gemini 2.5 Pro.} Grok 4 (48.4\%) and Gemini 2.5 Pro (45.4\%) both fail almost half the time due to overlapping mutually exclusive tasks, suggesting difficulty reasoning over exclusivity constraints.
    \item \textbf{Temporal errors are non-trivial across most models.} Claude 4 Sonnet (12.1\%), GPT-4o (11.2\%), GPT-5 (8.9\%) and Qwen3 (6.8\%) show a meaningful share of temporal window violations, pointing to challenges in enforcing deadlines and release times.
    \item \textbf{Mixed failure profiles in “Other.”} DeepSeek R1 has the highest proportion of “Other” errors (25.6\%), suggesting frequent compound or non-standard violations. Claude 4 Sonnet also shows a relatively high “Other” rate (12.1\%).
    \item \textbf{Resource/Downtime is rarely the dominant issue.} Although present in every model’s error profile, this category generally remains under 12\%, implying that most models can respect downtime constraints when they fail elsewhere.
\end{itemize}

Overall, these patterns reveal that while some models exhibit concentrated weaknesses (e.g., precedence ordering for o3), others distribute their errors across multiple constraint types. This breakdown is valuable for guiding targeted improvements.

\section{DISCUSSION AND LIMITATIONS}
The analysis reveals that even the strongest models—\textbf{o3}, \textbf{Grok~4}, and \textbf{GPT-5}—exhibit sharply divergent strengths depending on the constraint profile and domain. \textbf{o3} dominates on the synthetic MCI constraints track, sustaining feasibility deep into the difficulty ramp, yet it suffers the largest negative transfer in the \emph{Data Center Migration} setting. \textbf{Grok~4} shows the reverse pattern: modest synthetic performance but substantially better real-world transfer. \textbf{GPT-5} lifts the real-world ceiling further (best observed WAUC and latest real-world BP), while remaining competitive synthetically; however, its real-world feasibility still declines in the tail, indicating that late-level constraint density remains a bottleneck even for the strongest model. Taken together, these outcomes show that success in purely synthetic RCPSP benchmarks is \emph{necessary but not sufficient} for robustness in constraint-rich operations, and that domain grounding can materially alter the failure surface.

The infeasibility analysis further clarifies these differences. Models with high synthetic scores often fail predominantly on precedence violations, while those with stronger domain transfer suffer more from disjunctive or temporal window violations. These patterns indicate that constraint-specific reasoning weaknesses, rather than generalized “hardness,” drive many failures. This has implications for model selection: practitioners should match model strengths to the constraint types most prevalent in their operational context.

Interpreting these findings also requires recognizing the design boundaries of our study. First, we evaluate nine models under a single-shot prompting protocol; while this controls for prompt variance, it does not capture the effects of multi-step reasoning, tool integration, or self-verification strategies that could yield higher feasibility rates~\cite{cot,tot,react,pal}. Second, the instance generator emphasizes five-layer DAGs with sampled temporal, downtime, and disjunctive constraints at fixed probabilities—ideal for controlled scaling but narrower than the correlated, irregular patterns found in real deployments. Third, we frame scheduling as a \emph{feasibility} task, using CP-SAT for verification; this design isolates reasoning accuracy from optimization, yet leaves open the question of whether feasible outputs can also meet makespan or cost objectives without solver assistance. Future work could complement pass/fail feasibility with a \emph{feasibility distance} metric that quantifies how far a schedule is from validity—e.g., minimal slack required to satisfy constraints, or minimal start-time adjustments—to enable graded scoring. Fourth, our real-world translation focuses on data center migration; while representative of a demonstrative scheduling domain, other areas such as manufacturing, construction, or satellite operations may exhibit different constraint interactions and failure distributions. 

By integrating these limitations into the interpretation of results, we position R-ConstraintBench as both a reliable measure of current prompt-only LLM scheduling capability and a foundation for future exploration into constraint-aware and domain-adapted reasoning approaches.

\section{CONCLUSION}
Constraint-rich scheduling underpins operations across industries, yet prevailing evaluations rarely expose where LLM reasoning fails when heterogeneous rules must be satisfied simultaneously. R-ConstraintBench addresses this gap by isolating latent feasibility reasoning, ramping \emph{constraint interaction} and density (MCI), and testing transfer in a domain-grounded data-center instantiation.

Two results are decisive. First, \emph{constraint interaction—not graph depth—is the bottleneck}: strong models are near-ceiling on precedence-only DAGs, but feasibility collapses once downtime, temporal windows, and disjunctive exclusivity interact. Second, \emph{synthetic success does not guarantee domain transfer}: models that excel on clean synthetic ramps can degrade when constraints are coupled as they are in practice.

Across nine models, we observe steep drops when constraint types interact. While \textbf{o3} leads on the synthetic MCI track, \textbf{GPT-5} attains the strongest data center score (and latest breakpoint), and all systems decline in the highest-constraint band. Failure profiles are diagnostic: precedence and disjunctive violations dominate many errors, with temporal and resource/downtime violations appearing less frequently.

%% *** EDIT: speculative phrasing per advisor guidance
We speculate that training and evaluation regimes emphasizing constraint interaction—rather than depth alone—will improve performance on RCPSP-style tasks and may support broader reasoning generalization; testing this explicitly is left to future work.

\section{FUTURE WORK}
Several directions can extend and deepen the impact of R-
ConstraintBench:

\begin{itemize}
    \item \textbf{Finetuning for Scheduling Competence:} Explore whether finetuning models on RCPSP-style problems, with either synthetic or domain-specific schedules, improves constraint satisfaction rates and robustness under high-density conditions.
    \item \textbf{Ablations on Prompting and Instance Factors:} Systematically vary prompt formulation (instruction framing, schema strictness, verbosity, ordering, units), constraint mix (proportions and correlation of downtime/temporal/disjunctive), and structural scale (tasks per layer, number of layers, resource types, calendar granularity). Quantify sensitivity of WAUC, BP-level, feasibility-distance, and error taxonomy to these factors.
    \item \textbf{Expanding to Additional NP-Complete Problems:} Extend the benchmark to include other combinatorial optimization problems such as the Traveling Salesman Problem, Job-Shop Scheduling, and Vehicle Routing Problems, enabling broader assessment of reasoning and optimization capabilities.
    \item \textbf{Multi-Domain Evaluation:} Introduce new operational domains beyond data center migration (e.g., construction sequencing, supply chain coordination, satellite task scheduling) to evaluate transfer learning and domain adaptation.
    \item \textbf{Prompting Strategy Variations:} Systematically study the impact of chain-of-thought prompting, step-by-step reasoning scaffolds, verification prompts, and retrieval-augmented generation on schedule feasibility and constraint adherence.
    \item \textbf{Code-Generation for Solvers:} Given the growing ability of modern LLMs to produce and execute code, evaluate model performance in generating complete, correct CP-SAT solver programs for given scheduling problems, comparing direct-scheduling outputs with solver-assisted outputs.
\end{itemize}

\section{REFERENCES}
\vspace{-2em}
\renewcommand{\refname}{}

% --- APPENDIX: PROMPTS ---
\clearpage
\onecolumn
\appendix
\section*{APPENDIX}

% Prompt 1 (Using the robust 'listings' package)
\begin{tcolorbox}[title=Phase IIa: Multi-Constraint Interaction Sample Prompt]
\begin{lstlisting}[basicstyle=\ttfamily\small]
You are a project scheduler tasked with creating a feasible schedule for a resource-constrained project scheduling problem.


- Task_5: Duration 5 time units, Requires: Resource_2, Resource_3
- Task_2: Duration 8 time units, Requires: Resource_2, Resource_1, Cannot start before time 14
- Task_7: Duration 10 time units, Cannot start before time 3
- Task_4: Duration 4 time units, Depends on: Task_7, Cannot start before time 17, Must finish by time 53
- Task_6: Duration 7 time units, Depends on: Task_2, Task_5, Cannot start before time 4
- Task_8: Duration 3 time units, Depends on: Task_7, Task_5, Task_2, Cannot start before time 10
- Task_10: Duration 1 time units, Depends on: Task_2, Task_7, Task_4, Cannot start before time 15, Must finish by time 58
- Task_11: Duration 3 time units, Requires: Resource_1, Resource_3, Depends on: Task_7, Task_5, Task_8, Task_6, Cannot start before time 6
- Task_3: Duration 4 time units, Requires: Resource_2, Depends on: Task_7, Task_6, Task_5, Must finish by time 73
- Task_1: Duration 5 time units, Depends on: Task_10, Task_6, Task_8, Cannot start before time 8, Must finish by time 47
- Task_9: Duration 2 time units, Requires: Resource_1, Depends on: Task_6, Task_1, Must finish by time 63
- Task_12: Duration 4 time units, Requires: Resource_3, Depends on: Task_3, Task_4, Task_5, Task_6, Task_10

- Resource_1: Capacity 3 (can handle 3 tasks simultaneously), Downtime windows: unavailable 8-10
- Resource_2: Capacity 1 (can handle 1 tasks simultaneously), Downtime windows: unavailable 18-20
- Resource_3: Capacity 1 (can handle 1 tasks simultaneously), Downtime windows: unavailable 33-37

- Task_2 and Task_7 cannot run simultaneously
- Task_4 and Task_9 cannot run simultaneously
- Task_10 and Task_9 cannot run simultaneously
- Task_10 and Task_12 cannot run simultaneously
- Task_11 and Task_12 cannot run simultaneously
- Task_7 and Task_11 cannot run simultaneously
- Task_7 and Task_6 cannot run simultaneously
- Task_9 and Task_12 cannot run simultaneously
- Task_2 and Task_4 cannot run simultaneously
- Task_2 and Task_6 cannot run simultaneously
- Task_8 and Task_3 cannot run simultaneously
- Task_6 and Task_1 cannot run simultaneously
- Task_2 and Task_11 cannot run simultaneously
- Task_2 and Task_10 cannot run simultaneously
- Task_10 and Task_11 cannot run simultaneously
- Task_8 and Task_9 cannot run simultaneously
- Task_2 and Task_1 cannot run simultaneously
- Task_4 and Task_12 cannot run simultaneously
- Task_6 and Task_3 cannot run simultaneously
- Task_5 and Task_7 cannot run simultaneously
- Task_4 and Task_8 cannot run simultaneously
- Task_10 and Task_1 cannot run simultaneously
- Task_2 and Task_8 cannot run simultaneously
- Task_4 and Task_11 cannot run simultaneously
- Task_5 and Task_3 cannot run simultaneously
- Task_7 and Task_9 cannot run simultaneously
- Task_7 and Task_3 cannot run simultaneously
- Task_6 and Task_10 cannot run simultaneously
- Task_8 and Task_12 cannot run simultaneously
- Task_11 and Task_1 cannot run simultaneously
- Task_7 and Task_10 cannot run simultaneously
- Task_7 and Task_1 cannot run simultaneously
- Task_8 and Task_10 cannot run simultaneously
- Task_3 and Task_9 cannot run simultaneously
- Task_5 and Task_1 cannot run simultaneously
- Task_7 and Task_8 cannot run simultaneously
- Task_8 and Task_1 cannot run simultaneously
- Task_5 and Task_6 cannot run simultaneously
- Task_5 and Task_10 cannot run simultaneously
- Task_5 and Task_8 cannot run simultaneously
- Task_4 and Task_1 cannot run simultaneously
- Task_5 and Task_4 cannot run simultaneously
- Task_6 and Task_9 cannot run simultaneously
- Task_3 and Task_12 cannot run simultaneously
- Task_4 and Task_3 cannot run simultaneously
- Task_1 and Task_12 cannot run simultaneously
- Task_11 and Task_3 cannot run simultaneously
- Task_4 and Task_6 cannot run simultaneously
- Task_5 and Task_9 cannot run simultaneously

- Task_3 must finish before Task_12 can start
- Task_2 must finish before Task_10 can start
- Task_2 must finish before Task_6 can start
- Task_7 must finish before Task_3 can start
- Task_4 must finish before Task_12 can start
- Task_6 must finish before Task_9 can start
- Task_7 must finish before Task_8 can start
- Task_10 must finish before Task_1 can start
- Task_7 must finish before Task_11 can start
- Task_5 must finish before Task_12 can start
- Task_6 must finish before Task_1 can start
- Task_8 must finish before Task_1 can start
- Task_1 must finish before Task_9 can start
- Task_7 must finish before Task_10 can start
- Task_6 must finish before Task_12 can start
- Task_5 must finish before Task_8 can start
- Task_7 must finish before Task_4 can start
- Task_4 must finish before Task_10 can start
- Task_6 must finish before Task_3 can start
- Task_10 must finish before Task_12 can start
- Task_2 must finish before Task_8 can start
- Task_5 must finish before Task_11 can start
- Task_8 must finish before Task_11 can start
- Task_5 must finish before Task_3 can start
- Task_6 must finish before Task_11 can start
- Task_5 must finish before Task_6 can start

Create a feasible schedule that minimizes the total project duration (makespan) while respecting all constraints.

Please provide your solution in JSON format with the following structure:
{
  "makespan": <total_duration>,
  "schedule": [
    {
      "task": "<task_id>",
      "start_time": <start_time>,
      "end_time": <end_time>,
      "resources": ["<resource_id>"]
    }
  ]
}

\end{lstlisting}
\end{tcolorbox}

\clearpage

% Prompt 2 (Using the robust 'listings' package)
\begin{tcolorbox}[title=Phase IIb: Real-World Data Migration Sample Prompt]
\begin{lstlisting}[basicstyle=\ttfamily\small]
You are managing a critical data center migration from Facility 1 to Facility 2. This migration involves 2 server racks that must be carefully moved while minimizing total downtime.

MIGRATION OVERVIEW:
- Number of racks: 2
- Total tasks: 10
- Migration phases per rack: Shutdown → Unrack → Transport → Install → Test
- Complexity level: 1

MIGRATION TASKS:
Each rack must go through the following sequence:

Rack_1:
  - Rack_1_shutdown: 27 minutes, Requires: IT_Team
  - Rack_1_unrack: 40 minutes, Requires: DC_Crew, Forklift, Cannot start before minute 4
  - Rack_1_transport: 33 minutes, Requires: Convoy
  - Rack_1_install: 45 minutes, Requires: DC_Crew, Forklift
  - Rack_1_test: 32 minutes, Requires: IT_Team, Network_Engineers

Rack_2:
  - Rack_2_shutdown: 26 minutes, Requires: IT_Team
  - Rack_2_unrack: 20 minutes, Requires: DC_Crew, Forklift, Must complete by minute 52
  - Rack_2_transport: 58 minutes, Requires: Convoy, Cannot start before minute 32
  - Rack_2_install: 41 minutes, Requires: DC_Crew, Forklift, Cannot start before minute 43, Must complete by minute 300
  - Rack_2_test: 34 minutes, Requires: IT_Team, Network_Engineers

AVAILABLE RESOURCES:
- IT_Team: Can handle 2 tasks simultaneously
- DC_Crew: Can handle 3 tasks simultaneously
- Network_Engineers: Can handle 2 tasks simultaneously
- Forklift: Can handle 1 task at a time
- Convoy: Can handle 1 task at a time (Downtime: unavailable minutes 169-203, unavailable minutes 41-88)

CRITICAL INTER-RACK DEPENDENCIES:
The following tasks have dependencies across racks:
- Rack_2_shutdown must complete before Rack_1_test

EQUIPMENT CONFLICTS:
The following task pairs cannot run simultaneously due to shared equipment:
- Rack_1_install and Rack_2_unrack (both need Forklift, DC_Crew)
- Rack_1_install and Rack_2_install (both need Forklift, DC_Crew)
- Rack_1_install and Rack_1_unrack (both need Forklift, DC_Crew)
- Rack_1_unrack and Rack_2_install (both need Forklift, DC_Crew)
- Rack_1_unrack and Rack_2_unrack (both need Forklift, DC_Crew)

PRECEDENCE CONSTRAINTS:
- Rack_1_shutdown must finish before Rack_1_unrack can start
- Rack_1_unrack must finish before Rack_1_transport can start
- Rack_1_transport must finish before Rack_1_install can start
- Rack_1_install must finish before Rack_1_test can start
- Rack_2_shutdown must finish before Rack_2_unrack can start
- Rack_2_unrack must finish before Rack_2_transport can start
- Rack_2_transport must finish before Rack_2_install can start
- Rack_2_install must finish before Rack_2_test can start
- Rack_2_shutdown must finish before Rack_1_test can start

CONSTRAINTS:
1. Each rack's tasks must follow the sequence: Shutdown → Unrack → Transport → Install → Test
2. Tasks require specific crews and equipment with limited capacity
3. Resources have scheduled downtime windows (breaks, maintenance)
4. Some tasks have time windows (earliest start, latest finish)
5. Tasks sharing limited equipment cannot run simultaneously
6. Inter-rack dependencies must be respected

OBJECTIVE:
Create a migration schedule that completes all tasks in a feasible time while respecting all constraints.

Please provide your solution in JSON format with the following structure:
{
  "makespan": <total_duration>,
  "schedule": [
    {
      "task": "<task_id>",
      "start_time": <start_time>,
      "end_time": <end_time>,
      "resources": ["<resource_id>"]
    }
  ]
}
\end{lstlisting}
\end{tcolorbox}
\newpage

\begin{tcolorbox}[title=Error Analysis on Model Output Sample]
\begin{lstlisting}[basicstyle=\ttfamily\small]
Prompt: 
You are a project scheduler tasked with creating a feasible schedule for a resource-constrained project scheduling problem.

- Task_5: Duration 3 time units
- Task_3: Duration 7 time units, Requires: Resource_1, Cannot start before time 5, Must finish by time 16
- Task_1: Duration 2 time units, Requires: Resource_2, Depends on: Task_3, Cannot start before time 10
- Task_2: Duration 5 time units, Requires: Resource_3, Resource_1, Depends on: Task_5, Cannot start before time 6
- Task_4: Duration 6 time units, Depends on: Task_2, Task_1, Must finish by time 27
- Task_6: Duration 6 time units, Depends on: Task_3, Cannot start before time 3, Must finish by time 20
- Task_7: Duration 8 time units, Depends on: Task_5, Task_1, Cannot start before time 9
- Task_8: Duration 2 time units, Requires: Resource_2, Depends on: Task_5, Task_1, Task_4, Cannot start before time 4

- Resource_1: Capacity 1 (can handle 1 tasks simultaneously)
- Resource_2: Capacity 1 (can handle 1 tasks simultaneously), Downtime windows: unavailable 4-9
- Resource_3: Capacity 1 (can handle 1 tasks simultaneously), Downtime windows: unavailable 6-9

- Task_5 and Task_2 cannot run simultaneously
- Task_5 and Task_1 cannot run simultaneously
- Task_3 and Task_1 cannot run simultaneously
- Task_5 and Task_6 cannot run simultaneously
- Task_3 and Task_8 cannot run simultaneously
- Task_3 and Task_7 cannot run simultaneously
- Task_1 and Task_7 cannot run simultaneously
- Task_3 and Task_4 cannot run simultaneously
- Task_5 and Task_4 cannot run simultaneously
- Task_4 and Task_6 cannot run simultaneously
- Task_4 and Task_8 cannot run simultaneously
- Task_2 and Task_8 cannot run simultaneously
- Task_6 and Task_7 cannot run simultaneously
- Task_2 and Task_7 cannot run simultaneously
- Task_1 and Task_4 cannot run simultaneously
- Task_7 and Task_8 cannot run simultaneously
- Task_1 and Task_2 cannot run simultaneously
- Task_4 and Task_7 cannot run simultaneously
- Task_1 and Task_6 cannot run simultaneously
- Task_5 and Task_7 cannot run simultaneously
- Task_6 and Task_8 cannot run simultaneously

- Task_3 must finish before Task_1 can start
- Task_5 must finish before Task_2 can start
- Task_2 must finish before Task_4 can start
- Task_5 must finish before Task_7 can start
- Task_3 must finish before Task_6 can start
- Task_1 must finish before Task_4 can start
- Task_1 must finish before Task_7 can start
- Task_5 must finish before Task_8 can start
- Task_1 must finish before Task_8 can start
- Task_4 must finish before Task_8 can start

Create a feasible schedule that minimizes the total project duration (makespan) while respecting all constraints.

Please provide your solution in JSON format with the following structure:
{
  "makespan": <total_duration>,
  "schedule": [
    {
      "task": "<task_id>",
      "start_time": <start_time>,
      "end_time": <end_time>,
      "resources": ["<resource_id>"]
    }
  ]
}

Model Solution Attempt: 
{
  "makespan": 23,
  "schedule": [
    {"task": "Task_5", "start_time": 0, "end_time": 3, "resources": []},
    {"task": "Task_3", "start_time": 5, "end_time": 12, "resources": ["Resource_1"]},
    {"task": "Task_1", "start_time": 12, "end_time": 14, "resources": ["Resource_2"]},
    {"task": "Task_2", "start_time": 14, "end_time": 19, "resources": ["Resource_1", "Resource_3"]},
    {"task": "Task_6", "start_time": 12, "end_time": 18, "resources": ["Resource_1"]},
    {"task": "Task_4", "start_time": 19, "end_time": 25, "resources": []},
    {"task": "Task_7", "start_time": 14, "end_time": 22, "resources": []},
    {"task": "Task_8", "start_time": 25, "end_time": 27, "resources": ["Resource_2"]}
  ]
}

Infeasibility Analysis: 
• Resource capacity violation: Resource_1 has 2 tasks at time 14, capacity is 1
• Resource capacity violation: Resource_1 has 2 tasks at time 15, capacity is 1
• Resource capacity violation: Resource_1 has 2 tasks at time 16, capacity is 1
• Resource capacity violation: Resource_1 has 2 tasks at time 17, capacity is 1
• Disjunctive constraint violation: Task_6 and Task_7 overlap
• Disjunctive constraint violation: Task_2 and Task_7 overlap
• Disjunctive constraint violation: Task_4 and Task_7 overlap
• Disjunctive constraint violation: Task_1 and Task_6 overlap


\end{lstlisting}
\end{tcolorbox}

\end{document}